\documentclass{article}

\usepackage{microtype}
\usepackage{graphicx}
\usepackage{booktabs} 
\usepackage{Definitions}
\usepackage{xspace}
\usepackage{colortbl}
\usepackage[svgnames]{xcolor}
\usepackage{array,multirow,graphicx}
\usepackage{makecell}
\usepackage[utf8x]{inputenc}
\usepackage{textcomp}
\usepackage[ruled, linesnumbered]{algorithm2e}
\usepackage{hyperref}
\usepackage{cleveref}
\usepackage{subcaption}
\usepackage{multirow}

\SetKwComment{Comment}{$\triangleright$\ }{}
\newcommand{\skipl}{SkipLayer\xspace}

\usepackage[accepted]{icml2022}

\icmltitlerunning{Learning to Skip for Language Modeling}

\begin{document}

\twocolumn[
\icmltitle{Learning to Skip for Language Modeling}




\begin{icmlauthorlist}
\icmlauthor{Dewen Zeng}{former}
\icmlauthor{Nan Du}{former}
\icmlauthor{Tao Wang}{former}
\icmlauthor{Yuanzhong Xu}{google}
\icmlauthor{Tao Lei}{google}
\icmlauthor{Zhifeng Chen}{google}
\icmlauthor{Claire Cui}{google}
\end{icmlauthorlist}

\icmlaffiliation{former}{Work done when Dewen did his internship at Google Research and when Nan and Tao were at Google.}
\icmlaffiliation{google}{Google Brain}

\icmlcorrespondingauthor{Nan Du}{dunan@apple.com}
\icmlcorrespondingauthor{Zhifeng Chen}{zhifengc@google.com}

\icmlkeywords{Machine Learning, ICML}

\vskip 0.3in
]


\printAffiliationsAndNotice{}  

\begin{abstract}

Overparameterized large-scale language models have impressive generalization performance of in-context few-shot learning. However, most language models allocate the same amount of parameters or computation to each token, disregarding the complexity or importance of the input data. We argue that in language model pretraining, a variable amount of computation should be assigned to different tokens, and this can be efficiently achieved via a simple routing mechanism. Different from conventional early stopping techniques where tokens can early exit at only early layers, we propose a more general method that dynamically skips the execution of a layer (or module) for any input token with a binary router. In our extensive evaluation across 24 NLP tasks, we demonstrate that the proposed method can significantly improve the 1-shot performance compared to other competitive baselines only at mild extra cost for inference.


\end{abstract}

\section{Introduction}
\label{sec:intro}
Transformer-based~\cite{vaswani2017attention} large scale language models trained with general corpus have shown tremendously improvement of generalization in particular with in-context few-shot learning in recent years~\cite{shoeybi2019megatron, NEURIPS2020_gpt3, gopher2021, palm, hoffmann2022an}.
Despite of the impressive capability of text generation, training and serving these giant models are non-trivial even with the recent progress of hardware and software~\cite{jouppi2017datacenter,lepikhin2020gshard, patterson2021carbon}. One of the major challenges is that the processing of each input requires to activate all the parameters of a model, which often leads to trillions of floating point operations (FLOPs) per prediction. This imposes a big burden on both model training and inference since we have no control over the amount of computation that can be assigned to each input example.

In contrast, it is commonly believed that human cognition~\cite{stanovich_west_2000, Levy2008-LEVESC} uses varying cognitive efforts to operate and learn depending on the `hardness' of the input. Specifically, one may only need small efforts (lower computational cost) to process `easy' examples, like the commonly used stop-words, punctuation, patches in the background of an image, etc., but allow additional efforts (more computational cost) for `hard' examples, e.g., a rare abstract concept for reasoning, when they are truly needed. Therefore, allocating the same computational power of a large model uniformly for processing all samples tend to be wasteful and less efficient. Such issue might be even more exacerbated when training large models using real-world data corpus in that the redundancy of trivial examples will be more pronounced as more and more data are used.

Conditional computation \cite{BengioLC13, BengioBPP15} is the paradigm where only a small subset of the model parameters are activated based on the input representation, thereby reducing the amount of computation needed per example. However, due to the discreteness of the decisions based on each input, training neural networks with conditionally activated components end-to-end differentiably and efficiently is still challenging.

In this paper, we develop a simple framework, referred to as the~\skipl, which allows an input to skip any layer that can be wrapped inside it conditioned on the contextual representation. More specifically, ~\skipl-based models can be trained end-to-end differentiably while at the same time the discrete decisions during the forward pass can still be respected, which enables us to precisely control the performance-compute tradeoff through external constraint. Moreover, because the discrete decisions can be preserved during the forward pass, we also develop an efficient implementation so that the additional computation can be further saved in both pretraining and inference for the given target budget. We then apply~\skipl to the Transformer architecture~\cite{vaswani2017attention} to demonstrate the potential efficacy of the method for decoder-only language model pretraining and decoding. Finally, we extensively validate our method on a suite of well established NLP benchmarks ranging from open-domain QA tasks, reading comprehension, common sense reasoning, to natural language inference tasks. ~\skipl-based models have shown strong 1-shot performance with controllable computation tradeoff between model quality and decoding efficiency compared to a variety of competitive baselines.

\section{Method}
In this section, we elaborate on our proposed~\skipl framework, its efficient implementation, and the application to Transformer-based language models.

\subsection{SkipLayer}
\begin{figure}
    \centering
    \renewcommand\tabcolsep{4pt}
\begin{tabular}{cc}
\begin{subfigure}[b]{0.16\textwidth}
\includegraphics[width=\textwidth]{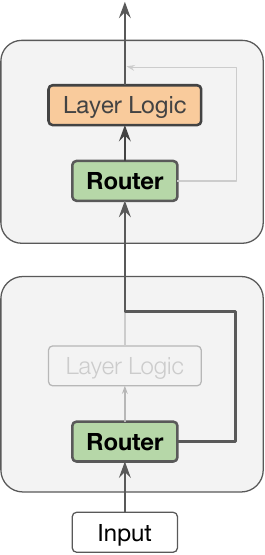}
\caption{}
\end{subfigure}
&
\begin{subfigure}[b]{0.31\textwidth}
\includegraphics[width=\textwidth]{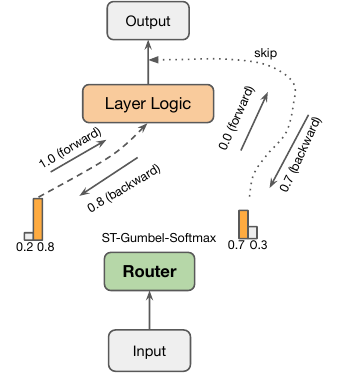}
\caption{}
\end{subfigure}
\end{tabular}
    \caption{(a) Overview of our SkipLayer framework. The router can choose to activate or skip the embeded layer logic based on the input context. (b) Straight-Through Gumbel-Softmax is used for the router. In the forward pass, binary variables are sampled. During the backward pass, gradients can be backpropagated to update the router.}
    \label{fig:SkipLayer}
\end{figure}

Let $X^{o}_{\text{[layer]}}=F_{[\text{layer}]}(X|\mathbb{W})$ denote a parameterized layer (or module) of a neural network with input $X$ and output $X^o$ given an optional set of weights represented by $\mathbb{W}$. For instance, a plain FeedForward layer (FFN) can be denoted as $X^{o}_{\text{FFN}}=F_{\text{FFN}}(X|\{W^i, W^o\})$ where $W^i\in\mathbb{R}^{m\times h}$ and $W^o\in\mathbb{R}^{h\times m}$ are the input and output weight, respectively.

A~\skipl $F_{\text{SL}}$ is designed to wrap an existing layer such that
\begin{align}
    X^{o}_{\text{SL}} & =F_{\text{SL}}\left(F_{[\text{layer}]}(X)|W_{\text{G}}\right)\\
    & = F_{[\text{layer}]}(X) \odot G(X|W_G) + X\odot (1 - G(X|W_G)),\nonumber
\end{align}
where $G(X|W_G)\in\{0, 1\}$ is a router function with the learnable weight $W_G$. Figure \ref{fig:SkipLayer}(a) shows the overall framework.

Given a batch $X\in\mathbb{R}^{B\times T\times d}$ of $B$ sequences, each of length $T$ and the embedding dimension $d$, for each token input $X[b, t]\in\mathbb{R}^d, b\leq B, t\leq T$, we have that
\begin{align}
    X^{o}_{\text{SL}}[b, t]=
    \begin{cases}
        F_{[\text{layer}]}\left(X[b, t]\right), & \text{if $G(X[b,t])=1$}.\\
        X[b,t], & \text{otherwise}.
    \end{cases}
    \label{eq:sl}
\end{align}

Therefore, as shown in Figure \ref{fig:SkipLayer}(a), any existing layer applied to the input in a pointwise way (e.g., FFN) can be easily embedded inside a~\skipl. Based on the context, if the router decides to skip, the input will be connected directly to the output, otherwise it will go through the embedded layer logic.

\paragraph{Router Function.}
Central to the~\skipl is the router function $G(X|W_G)$ which is learned to assign only a subset of inputs to the embedded layer for the best model performance under a given budget. For a batch of input tokens $X\in\mathbb{R}^{B\times T\times d}$, the router outputs a binary mask matrix
\begin{align}
M=G(X|W_G)\in\{0, 1\}^{B\times T}, W_G\in\mathbb{R}^{d\times 2}.
\label{eq:mask}
\end{align}
There are several choices for designing $G(X|W_G)$. One choice is the Sigmoid function $G(X|W_G)=\sigma (XW_G)$ that independently normalizes each value to be within the continuous range $(0, 1)$ as the soft approximation to the binary masking. Although this approximation is easy to differentiate, it needs an additional threshold to produce the binary decision rule of Equation~\ref{eq:sl}. 

The second design choice is the Top-K $(K=1)$ routing which is widely used in the works of \cite{du2022glam, lepikhin2020gshard}, giving $G(X|W_G)=\text{Top-1}\left(XW_G\right)=\argmax{XW_G}$. In order to address the indifferentiability of the $\argmax$ operator, as discussed in \cite{lepikhin2020gshard}, for each input token $X[b, t]$, we first normalize the dot-product scores by $g= \text{Softmax}(X[b, t]W_G)\in\mathbb{R}^2$, and let 
\begin{align}
    X^{o}_{\text{SL}}[b, t]=
    \begin{cases}
        g[1]\cdot F_{[\text{layer}]}\left(X[b, t]\right), & \text{if $\argmax g=1$}.\\
        g[0]\cdot X[b,t], & \text{otherwise},
    \end{cases}
    \label{eq:topk}
\end{align}
such that the gradients can be backpropgated through the coefficients $g$.
Unfortunately, in our experiments, we find that the Top-1 formulation cannot precisely control the sparsity of the model (which is crucial to the efficiency) since $g$ is still a soft approximation of the binary decision rule.

We thus formulate the router with the Straight-Through Gumbel-Softmax trick shown in Figure~\ref{fig:SkipLayer}(b). In the forward pass, sampled binary values are returned for the gating $G(X[b, t])$ as in Equation~\ref{eq:sl}. In the backward pass, the soft probabilities are used as $g$ in Equation~\ref{eq:topk} for the gradients to be propagated back to update the router weights. Because during the forward pass we are able to directly calculate the percentage of the tokens that are not skipped based on the binary masking of Equation~\ref{eq:mask}, we can better control the density of the model.




\paragraph{Router Capacity.}
\label{section:aux_loss}
The binary mask of the router output in Equation~\ref{eq:mask} is the assignment of a subset of tokens in a batch to the embedded layer inside a~\skipl. For simplicity, suppose each sequence in a batch of size $B$ has the same sequence length $T$. Then, the ratio $r=\frac{\sum_{i,j}M[i, j]}{B\times T}$ is the percentage (or probability) that a token is assigned to the layer, which is also referred to as the capacity. Consider $P$ as a global budget of how many input tokens can be assigned to a layer. Following~\cite{du2022glam, lepikhin2020gshard}, we introduce an auxiliary loss term $\ell_{\text{aux}} = \sum_i^L(r_i - P)^2$ where $r_i$ is the capacity of layer $i\leq L$, so that each layer will respect the budget constraint. The overall loss function of the model will be $\mathcal{L} = \ell_{\text{nll}} + \lambda\cdot\ell_{\text{aux}}$ where $\ell_{\text{nll}}$ is the negative log-likelihood of predicting the next token on average. By optimizing $\mathcal{L}$, on the one hand, the layer capacity will be pushed to be closer to the target probability $P$. On the other hand, the $\ell_{\text{aux}}$ term will continuously improve the model's predictive accuracy. Since the $\ell_{\text{aux}}$ term will enforce only $P$ percent of tokens in a batch to go with the layer, in order to reduce the first term $\ell_{\text{nll}}$, `hard' examples that lead to large marginal reduction on average will be prioritized while `easy' examples that already achieve low perplexity will be skipped in order to save FLOPs.

The router capacity enables the flexibility of controlling the performance-computation trade-off. Specifically, we can increase the number of layers in total while at the same time reduce target probability $P$ to keep the average number of activated layers roughly the same. This effectively separates the increase of model capacity from the computation cost per prediction, and makes it possible to trade off the increased model capacity for better prediction. During serving, we are then able to load the model that can best utilize the accelerators' memory, lead to the highest prediction quality, and only mildly increase the computation cost while still meeting the latency requirement simultaneously.





\begin{figure}
    \centering
    \includegraphics[clip, trim=110pt 35pt 110pt 40pt, width=1.0\linewidth]{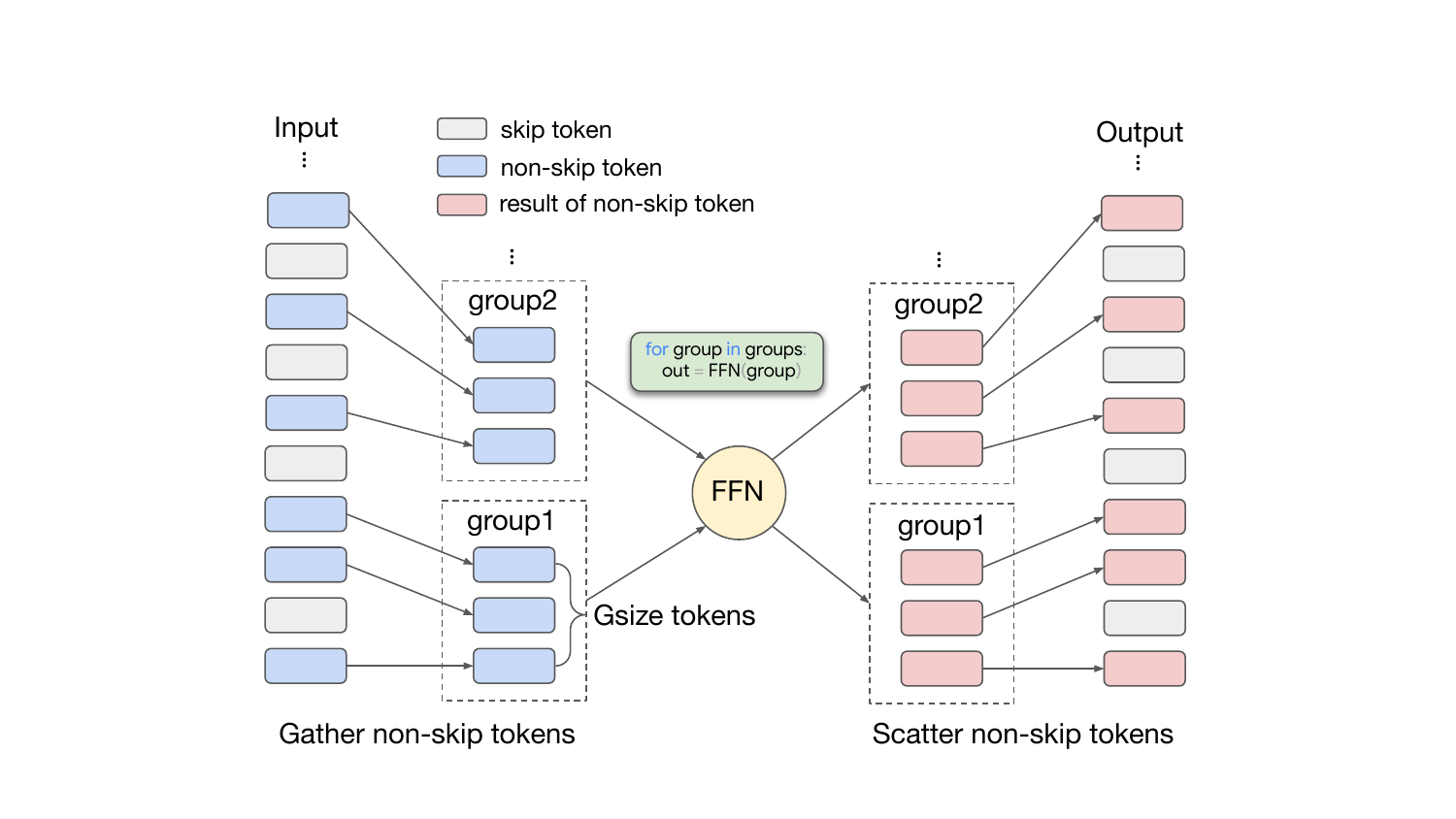}
    \caption{Illustration of efficient SkipLayer implementation. We focus on the sparse computation of FFN, non-skip tokens are gathered based on indices generated by the router and then fed into the FFN as groups. Gsize is a hyper-parameter that controls the group size. The results are scattered to the final output.}
    \label{fig:dynamic_gather_scatter}
\end{figure}

\begin{figure}[t]
    \centering
    \includegraphics[width=0.25\textwidth]{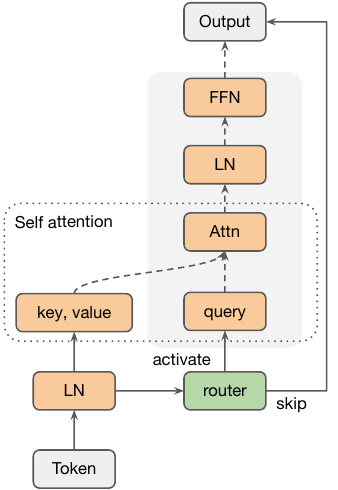}
    \caption{Overview of our \skipl for Transformer-based models, example of a single layer. LN is the layer normalization layer, query, key and value refers to the computation of query, key and value projections in the self-attention layer. Attn is the attention computation. Residual connections in the self attention layer and FFN layer are ignored for simplicity.}
    \label{fig:skiplayer_transformer}
\end{figure}

\subsection{Efficient Implementation}
\label{sec:efficition_impl}
The major advantage of~\skipl-based models is that the number of inputs computed by each layer is different across the entire stack of layers and continuously varies during training. At the same time, this dynamic characteristic is also challenging for implementation on TPU where computations of tensors with static shapes are often preferred. The basic implementation is to first apply the given layer logic in Figure~\ref{fig:SkipLayer}(a) to the entire batch and then multiply the output batch with the mask given by Equation~\ref{eq:mask} so that the skipped tokens will not be used in the layer. However, this masking mechanism is computational expensive since we should not spend the same computations on the skipped inputs as those on the non-skipped ones especially when the skip ratio is high.

We thus develop an efficient~\skipl implementation based on dynamic gather and scatter. The idea is illustrated in Figure \ref{fig:dynamic_gather_scatter} where we focus on the sparse computation of a FFN layer since it is a widely used component and often computationally intensive. The overall algorithm includes three major steps. 
\begin{enumerate}
    \item All inputs are marked as skip or non-skip based on results of the router in Equation~\ref{eq:mask}.
    \item All non-skipped inputs (in the blue rectangles) are gathered and evenly partitioned into groups. Although each group will be fed into the FFN for computation sequentially, all the elements in the same group will be gathered, computed, and scattered in parallel.
    \item The compute results from the non-skipped inputs will be scattered back to the final outputs of this FFN layer, while the skipped inputs will be directly written into the final outputs without any computation.
\end{enumerate}

The group size (the number of inputs in a group), denoted as Gsize, is a hyper-parameter that controls how many tokens will be processed by the FFN in parallel. Because the number of non-skipped inputs in a batch is dynamic and unknown in advance, Gsize affects the training efficiency. When Gsize is too large, e.g., there is only one single group, this group may include too many skipped inputs, leading to sub-optimal performance. When Gsize is too small, it will produce too many groups of small size, and the computation will be close to being sequential. Thus, there will be little parallelism, and the overheads maybe even larger than the basic masking implementation. In practice, we often set $\text{Gsize}\propto P\cdot BT$ where $P$ is the target probability, $B$ is the batch size, and $T$ is the sequence length.


\begin{algorithm}[t]
\small
\caption{Forward pass of \skipl}\label{alg:skiplayer_training}
\KwData{A batch of tokens $X\in\mathbb{R}^{B\times T\times d}$, target probability $P$.}
Get the mask $M$ by Equation~\ref{eq:mask}.\;

Get the key and value projection $K\leftarrow F_{\text{key}}(X)$, $V\leftarrow F_{\text{val}}(X)$.

\For{$b\leq B$, $t\leq T$}
{
\eIf{$M[b, t] = 1$}
{
    Get the query projection $q\leftarrow F_{\text{query}}(X[b, t])$
    
    $x^\prime\leftarrow F_{\text{Attn}}\left(F_{\text{LN}}(X[b, t])|K, q, V\right)\cdot M[b, t] + X[b, t]$
    
    $X_{\text{SL}}[b, t]\leftarrow F_{\text{FFN}}\left(F_{\text{LN}}(x^\prime)\right) + x^\prime$
}
{
    $X_{\text{SL}}[b, t]\leftarrow X[b, t])\cdot (1 - M[b, t])$
}
}
$\ell_{\text{aux}}\leftarrow (\sum_{b,t} M[b, t]/(B\cdot T) - P)^2$

\Return{$X_{\text{SL}}, \ell_{\text{aux}}$}
\end{algorithm}

\subsection{SkipLayer for Transformer-based Models}
In this section, we focus in particular on applying~\skipl to Transformer-based decoder-only language models in the setup of in-context learning.
A Transformer layer mainly includes the self-attention, layer normalization, and FFN as the sub-layers, and can be represented as
\begin{align}
    {X^l}^\prime &= F_{\text{Attn}}\left(F_{\text{LN}}(X^l)\right) + X^l, \nonumber\\
    X^{l+1} &= F_{\text{FFN}}\left(F_{\text{LN}}({X^l}^\prime)\right) + {X^l}^\prime.
\end{align}
\skipl can be applied to a single Transformer layer shown in ~Figure~\ref{fig:skiplayer_transformer}. We propose to wrap the entire Transformer layer into a~\skipl to preserve the atomicity of the self-attention ($F_{\text{Attn}}$) $\to$ FFN ($F_{\text{FFN}}$) structure. However, the $F_{\text{Attn}}$ layer and the $F_{\text{FFN}}$ layer have slightly different skipping implementations. 
The $F_{\text{FFN}}$ layer is often the most computationally intensive component of a Transformer model, but can be applied to a batch of tokens in a point-wise manner. Therefore, each input token of a batch can activate the $F_{\text{FFN}}$ layer independently with the probability $P$. Because $F_{\text{FFN}}$ consumes most of the computation in a Transformer layer, we can thus have big savings in FLOPs when the activation probability $P$ is small. 

The self-attention layer $F_{\text{Attn}}$ consumes much less computation relative to the $F_{\text{FFN}}$ layer. However, $F_{\text{Attn}}$ cannot apply to a batch of tokens in the pointwise way because tokens need to attend to each other to compute their own attention output. If most tokens in a batch are skipped when $P$ is small, the left non-skipped tokens will lose most of the context of the respective sequences they belong to. We also empirically observe lower predictive quality when this simple skipping mechanism is applied. Therefore, we propose the following partial skipping mechanism. As shown in Figure~\ref{fig:skiplayer_transformer}, when the input tokens are skipped, their key and value projections are still preserved (Line 2 in Algorithm \ref{alg:skiplayer_training}) since they are part of the context and are needed for the rest non-skipped tokens to further attend to. However, because the skipped tokens do not require attention calculations by themselves, we can still omit their query projections. It is worth to mention that Line 3-11 in Algorithm \ref{alg:skiplayer_training} can be computed in parallel using our efficient implementation in Section~\ref{sec:efficition_impl} to increase training speed.

Algorithm \ref{alg:skiplayer_decoding} shows the greedy decoding logic of one \skipl-based Transformer layer. The router makes the skipping decision by picking the most likely outcome. The key and value projections will be computed and saved in the decoding cache (Line 2). Only when the router activates the current layer, the query projection will be computed for current token, and then the decoding cache $K$ and $V$ which contain the key and value projections of previous decoding steps will be used to compute the self attention. Otherwise, there will be no further computations from this layer.

\begin{algorithm}
\small
\caption{\skipl per decoding step}\label{alg:skiplayer_decoding}
\KwData{The current state $x\in\mathbb{R}^d$, key and value cache $K$, $V$}
\setcounter{AlgoLine}{0}
$m = G(x|W_G)=\argmax{x^\top W_G}$

$K\leftarrow F_{\text{key}}(x)$, $V\leftarrow F_{\text{val}}(x)$

\eIf{$m = 1$}
{
    Get the query projection $q\leftarrow F_{\text{query}}(x)$
    
    $x^\prime\leftarrow F_{\text{Attn}}\left(F_{\text{LN}}(x)|K, q, V\right) + x$
    
    $x_{\text{SL}}\leftarrow F_{\text{FFN}}\left(F_{\text{LN}}(x^\prime)\right) + x^\prime$
}
{
    $x_{\text{SL}}\leftarrow x$
}

\textbf{return:} $x_{\text{SL}}$
\end{algorithm}

\section{Experiment Setup}
We focus on training decoder-only language models. This section elaborates our training setup, hyperparameters, baselines, benchmarks, and evaluation protocol.


\begin{table}[t]
\centering
\caption{Architectures and sizes of the models trained in our experiments. All trained model share the same learning hyperparameters.}
\begin{tabular}{llllc}
\toprule
Model      & $n_{\text{params}}$ & $P$ &  $L$ & Eff-L \\ \midrule
Standard (6L)     & 408M      & 0     & 6 & \multirow{4}{*}{6} \\
SkipLayer (12, 50\%) & 766M      & 50\%   & 12 & \\
SkipLayer (24, 25\%) & 1.47B      & 25\%   & 24 & \\
SkipLayer (48, 12.5\%) & 2.92B      & 12.5\%   & 48 & \\
\midrule
Standard (12L)     & 766M      & 0     & 12  & \multirow{4}{*}{12} \\
SkipLayer (24, 50\%) & 1.47B      & 50\%   & 24 & \\
SkipLayer (48, 25\%) & 2.92B      & 25\%   & 48 & \\
SkipLayer (96, 12.5\%) & 5.79B      & 12.5\%   & 96 & \\
\midrule
Standard (24L)     & 1.47B      & 0     & 24  & \multirow{3}{*}{24} \\
SkipLayer (48, 50\%) & 2.92B      & 50\%   & 48 & \\
SkipLayer (96, 25\%) & 5.79B      & 25\%   & 96 & \\
\bottomrule
\end{tabular}
\label{table:model_setup}
\end{table}


\paragraph{Dataset.}
The pretraining dataset has 1.6 trillion tokens that are representative of a wide range of natural language use cases. An in-house classifier is trained to classify between a collection of curated text and other web pages so that we are able to estimate the content quality of a webpage. A high-quality filtered subset of webpages are combined with books, Wikipedia pages, conversations, forums, and news to create the final dataset which is the same as~\cite{du2022glam} for training.

\begin{figure*}[tb]
\centering
    \renewcommand\tabcolsep{1pt}
\begin{tabular}{cccc}
\begin{subfigure}[b]{0.25\textwidth}
\includegraphics[width=\textwidth]{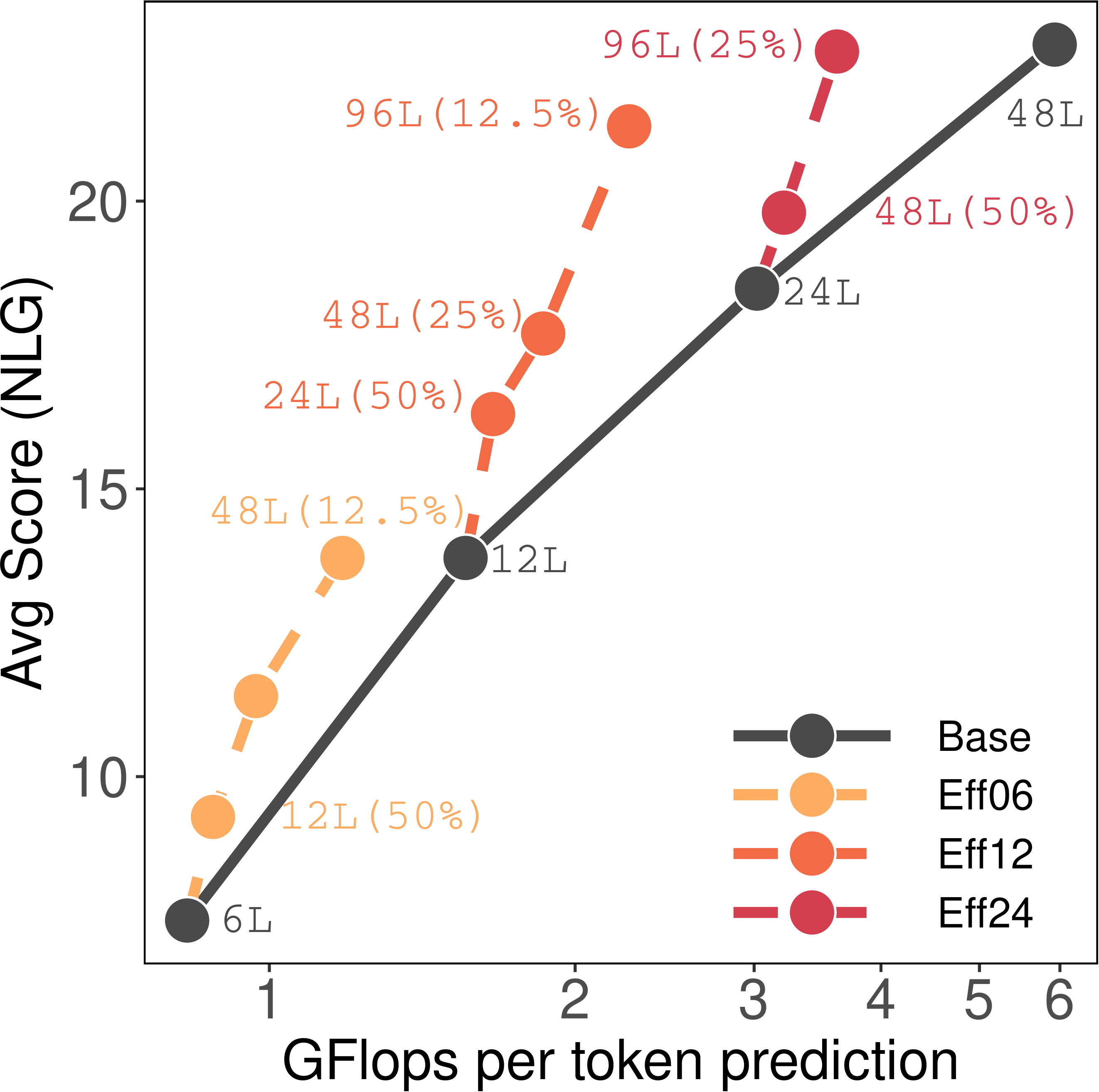}
\caption{Avg. 1-shot (NLG)}
\end{subfigure}
&
\begin{subfigure}[b]{0.25\textwidth}
\includegraphics[width=\textwidth]{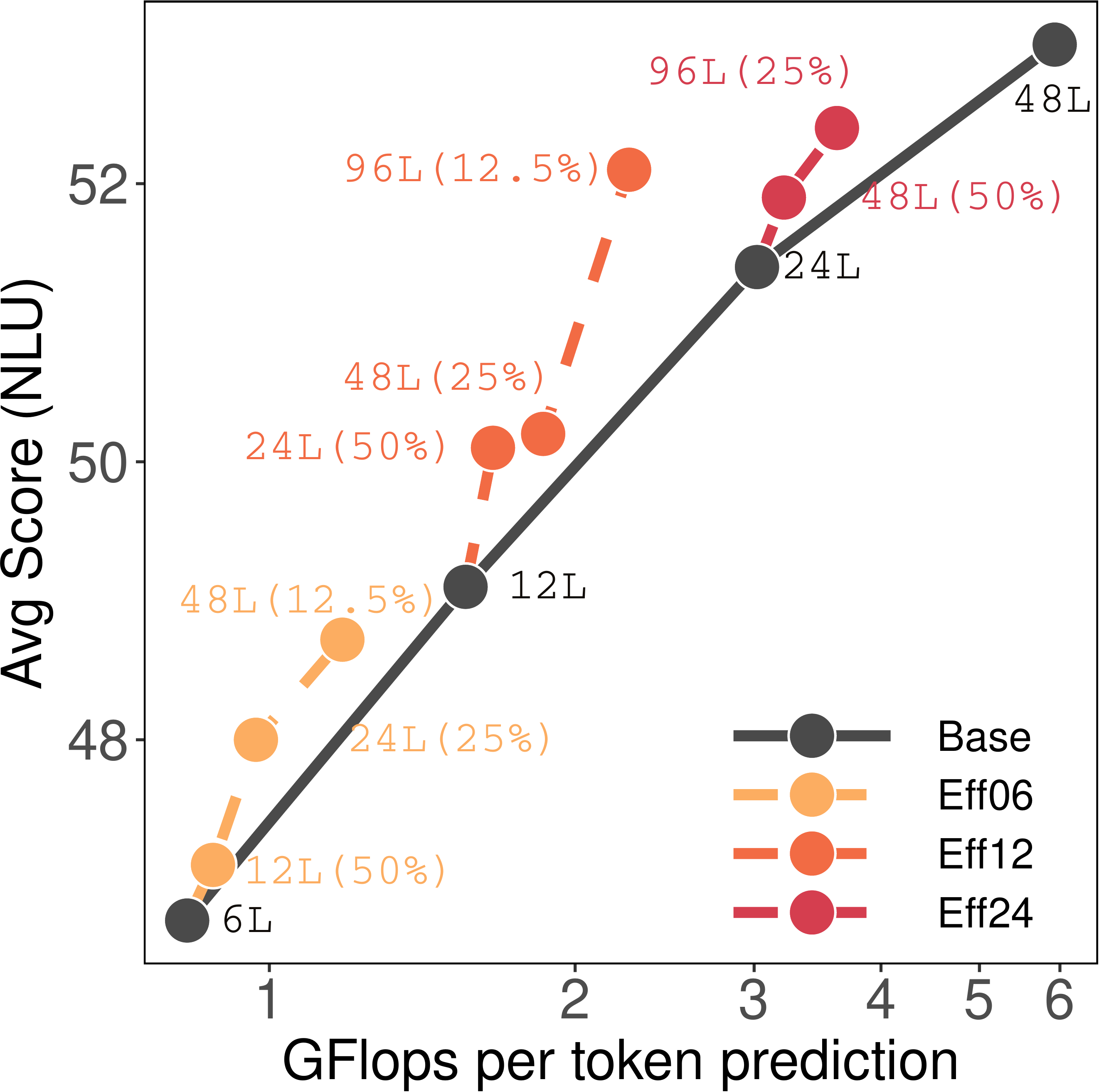}
\caption{Avg. 1-shot (NLU)}
\end{subfigure}
&
\begin{subfigure}[b]{0.25\textwidth}
\includegraphics[width=\textwidth]{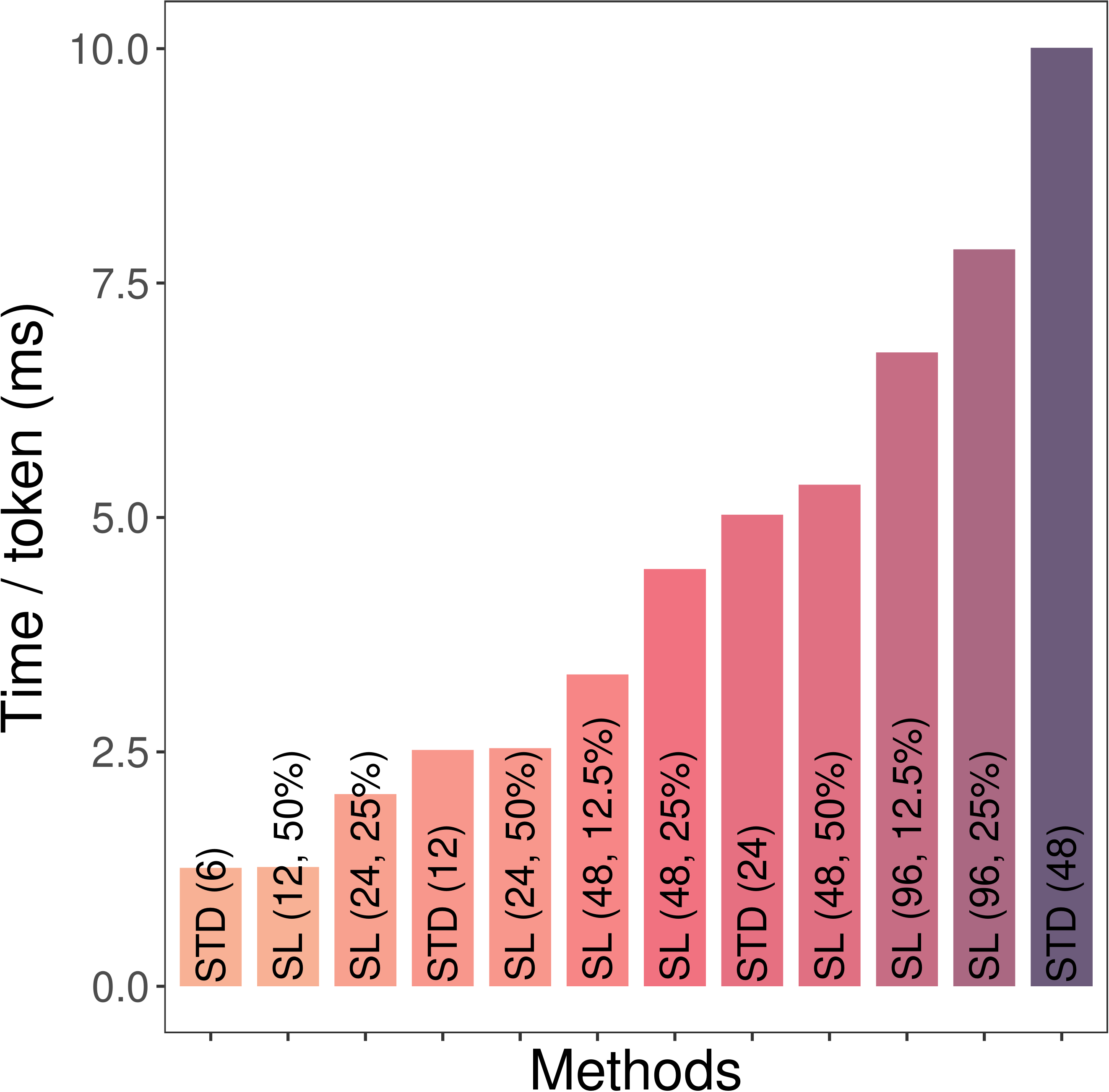}
\caption{Decoding time per token.}
\end{subfigure}
&
\begin{subfigure}[b]{0.25\textwidth}
\includegraphics[width=\textwidth]{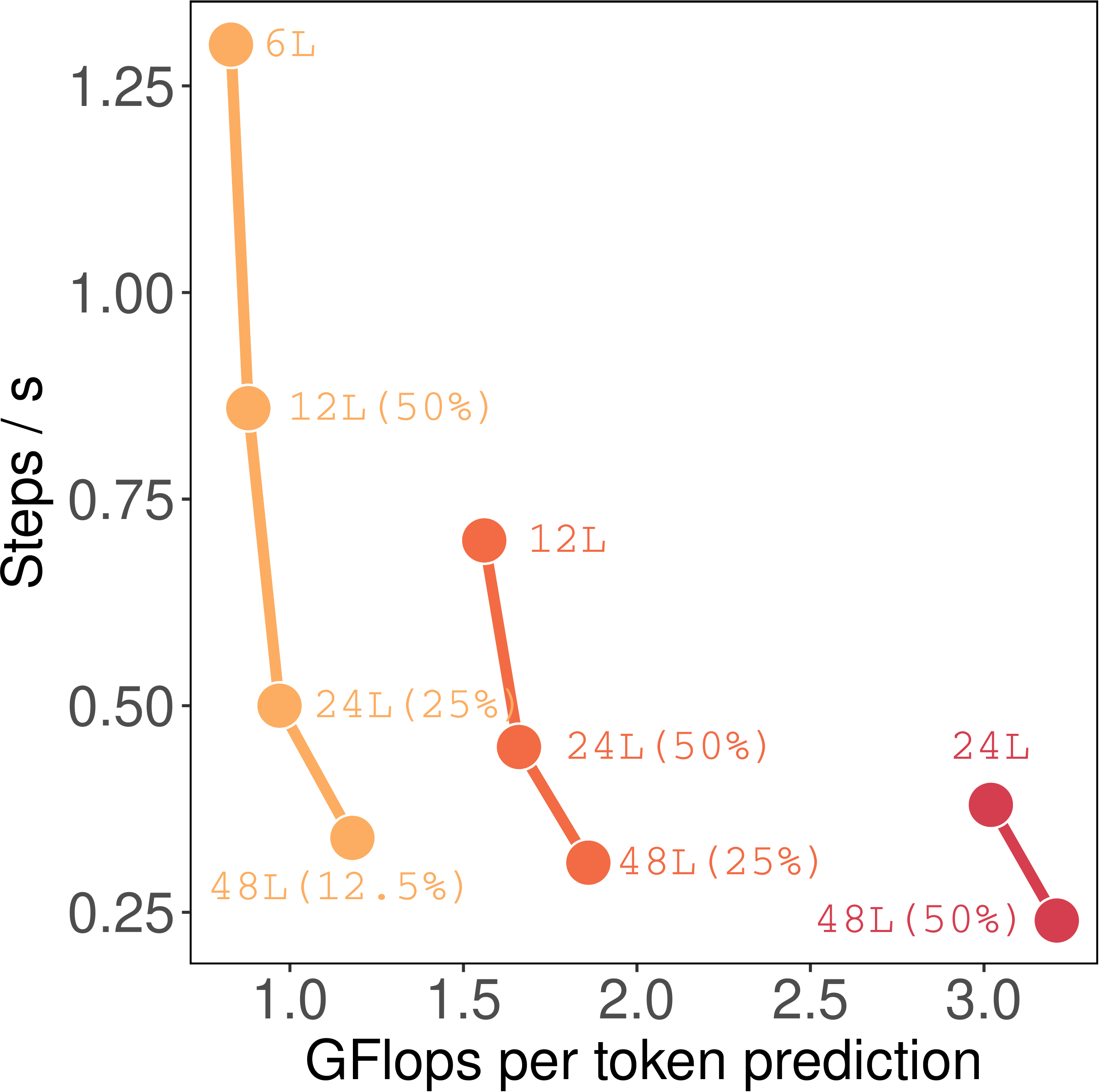}
\caption{Training speed.}
\end{subfigure}
\end{tabular}
\caption{Average 1-shot performance of different SkipLayer-based models for comparable effective FLOPs per token prediction over the NLG tasks (a) and NLU tasks (b).  (c) Comparisons of the decoding time per token between the SkipLayer-based models (SL) and the respective baseline models (STD). (d) Comparisons of the training speed among models of different density.}
\label{fig:performance}
\end{figure*}

\paragraph{Model Training.}
We have trained several variants of~\skipl-based models and baselines shown in Table~\ref{table:model_setup}. The model dimension of all the models is 1,536 and the hidden dimension of the FFN has 8$\times$ of the model dimension. The hidden dim of each attention head is 64. $n_{\text{params}}$ is the total number of trainable model parameters, $L$ is the total number of Transformer layers, $P$ is the probability of activating a layer, and Eff-L is the {\textbf{eff}}ective number of layers activated on average.
The sequence length is set to 1,024 tokens during training, and the batch size includes 256 sequences.
We set the learning rate to 0.1 for the first 10K training steps and then decay following an inverse square root schedule.
We use Adafactor optimizer with first-moment decay
$\beta_1 = 0$ and second-moment decay $\beta_2 = 0.99$.
The dropout rate is set to 0 as the processed tokens are token from an extremely large training corpus.
We use the SentencePiece \cite{Kudo2018SentencePieceAS} subword tokenizer with a vocabulary of size of 32K. 
During training, we use float32 for model weights and bfloat16 for activations.
Aside from the general negative log-likelihood loss, we add the auxiliary loss discussed in Section \ref{section:aux_loss} to control the skip ratio of our SkipLayer model, the auxiliary loss weight $\lambda$ is set to 0.1.
Finally, Gsize is set to 1,024 for all~\skipl-based models.

\paragraph{Model Evaluation.}
To directly evaluate the effectiveness of~\skipl-based models, we mainly follow the 1-shot learning protocol suggested by \cite{radford2019language}, which is widely used for evaluating the generalization quality of pre-trained language models. We evaluate each example in the development set of a benchmark. For each benchmark, only one example will be randomly drawn from that task's training set as the only demonstration and context, which will be then concatenated with the evaluation example with two newlines in between, and then fed into the model. We use exactly the same prompting format as \cite{radford2019language} for each downstream benchmark.

\paragraph{Benchmarks.}
We use 24 datasets including four natural language generative (NLG) tasks and 20 natural language understanding (NLU) tasks for evaluations. 
For NLG tasks, we compare the decoded sequence of tokens by the models to the ground-truth and report the Exact Match (EM) accuracy. These tasks are TriviaQA, NQS, WebQS, and SQuADv2. Greedy decoding is used for each task.
All NLU tasks are formulated into the form of selecting one correct answer from multiple candidate options. The prediction is based on the maximum log-likelihood of each option given the context $\log{P(\text{option}|\text{context})}$ normalized by the token length of each option. These NLU tasks include ANLI (R1, R2, R3), ARC (Easy, Challenge), BoolQ, CB, COPA, Hellaswag, Openbookqa, PIQA, Race (middle, high), ReCord, RTE, Storycloze, WIC, Winograd, Winogrande and WSC273.
Finally, we use the average of the scores across all datasets to report the overall 1-shot performance of models on both NLG and NLU tasks. 

\paragraph{Baselines.}
We consider the following baselines to study the effectiveness of \skipl-based models.
\begin{itemize}
    \item \textbf{Standard base model (STD).} The standard Transformer dense models without any skipping operations.
    \item \textbf{WideFFN.} Because the FFN often consumes the major computation and has big impact to the predictive performance~\cite{kocsis2022lowdataregime}, WideFFN is thus designed to further double the hidden dimension of the FFN component. We then apply~\skipl only to the FFN component without changing the total number of layers. As a consequence, when the skipping probability $P=50\%$, the compute FLOPs per prediction does not change due to the 2x FFN size.
    \item \textbf{HighwayNet}~\cite{srivastava2015highway} is among the first few works that propose to learn a gating function that could help to train very deep networks efficiently. We also apply this idea to the FFN component of the Transformer layer as one baseline.
    \item \textbf{Random Gating (Rand)}. Random gating method is the baseline where the learned gating function of a~\skipl model is replaced by a pure random function without learning. This baseline is designed to evaluate the importance of the learned gating function to a model's predictive performance.
\end{itemize}

\begin{figure*}[tb]
\centering
    \renewcommand\tabcolsep{1pt}
\begin{tabular}{cccc}
\begin{subfigure}[b]{0.25\textwidth}
\includegraphics[width=\textwidth]{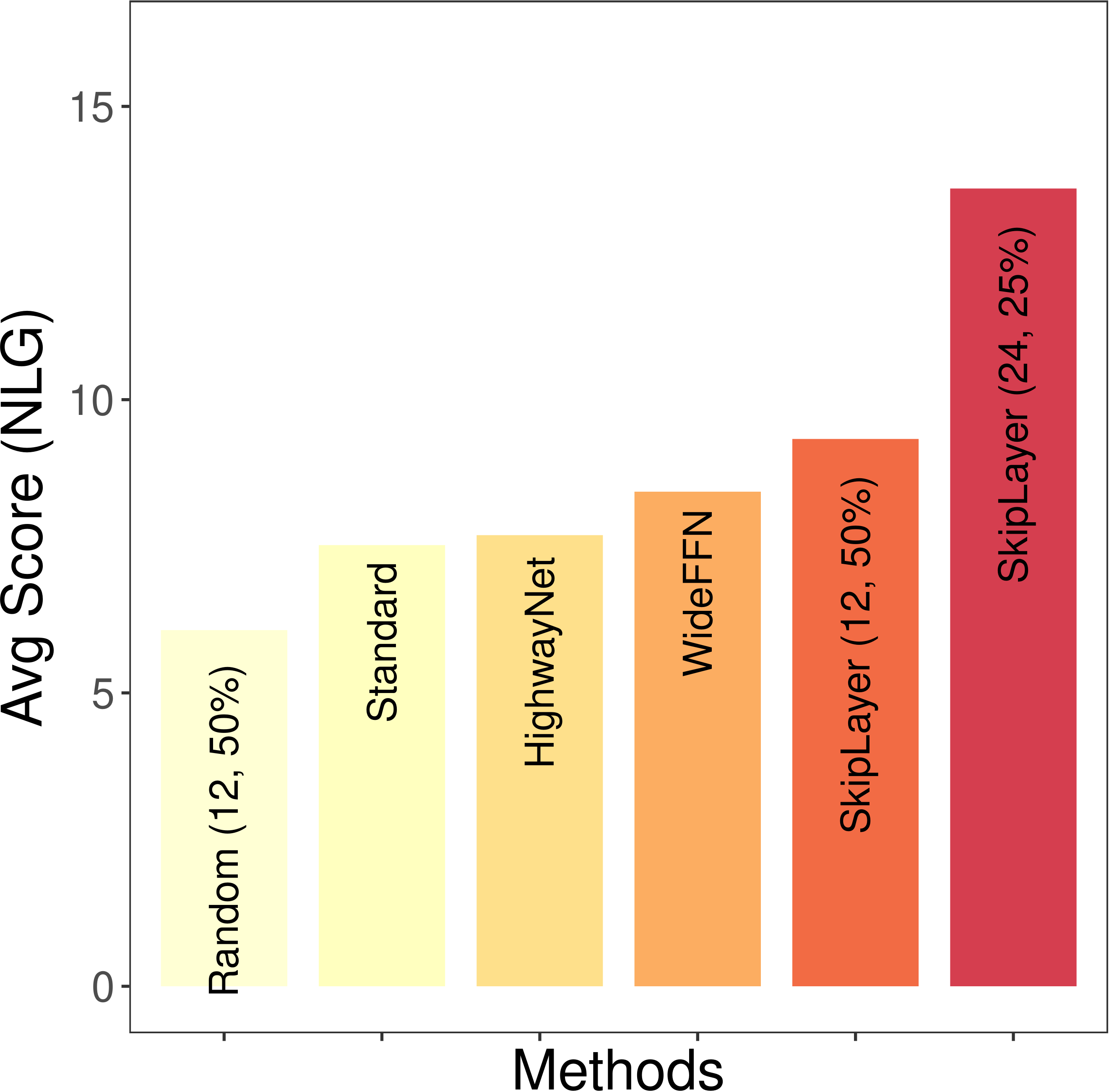}
\caption{Avg. 1-shot (NLG, 6L)}
\end{subfigure}
&
\begin{subfigure}[b]{0.25\textwidth}
\includegraphics[width=\textwidth]{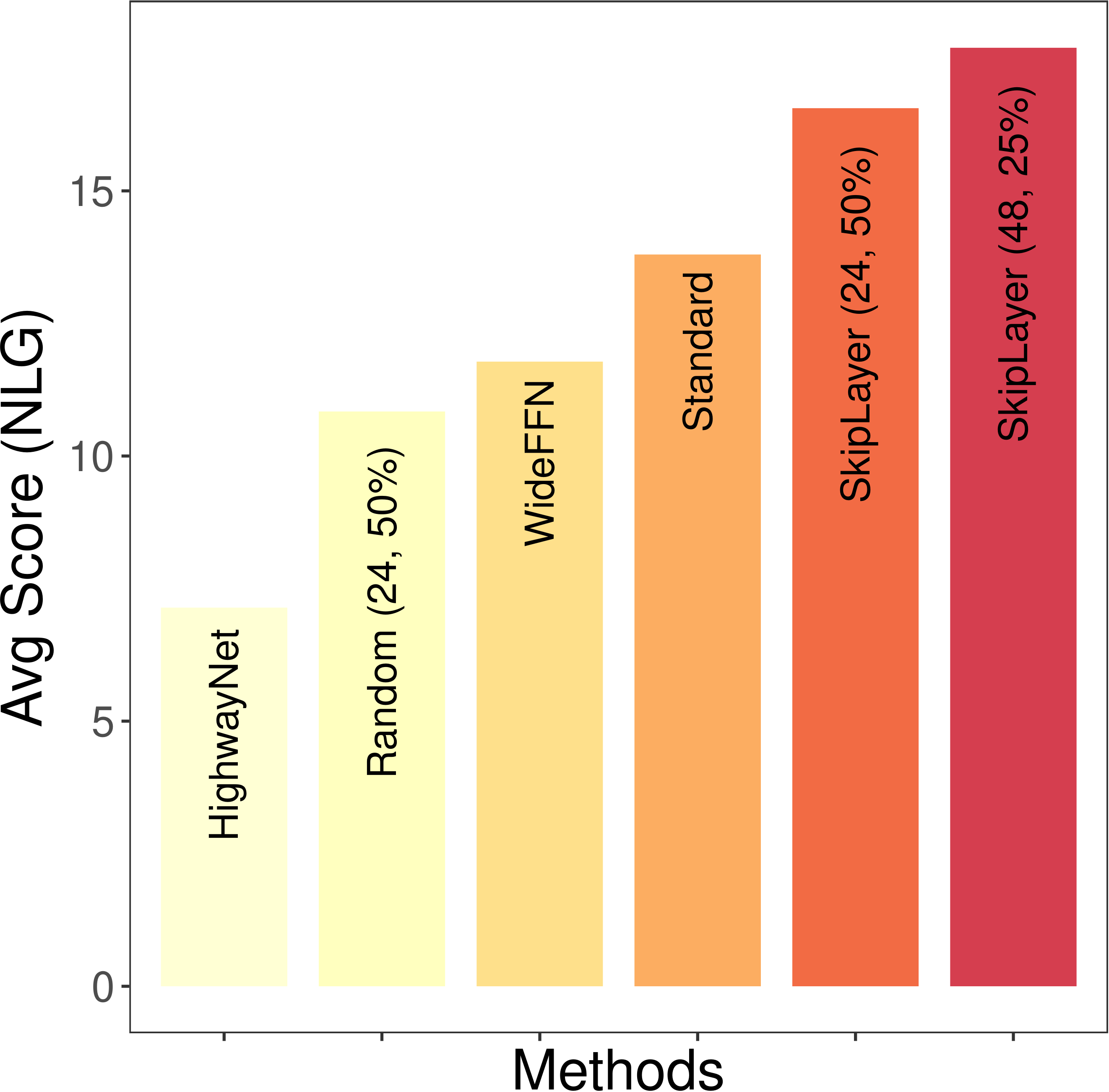}
\caption{Avg. 1-shot (NLG, 12L)}
\end{subfigure}
&
\begin{subfigure}[b]{0.25\textwidth}
\includegraphics[width=\textwidth]{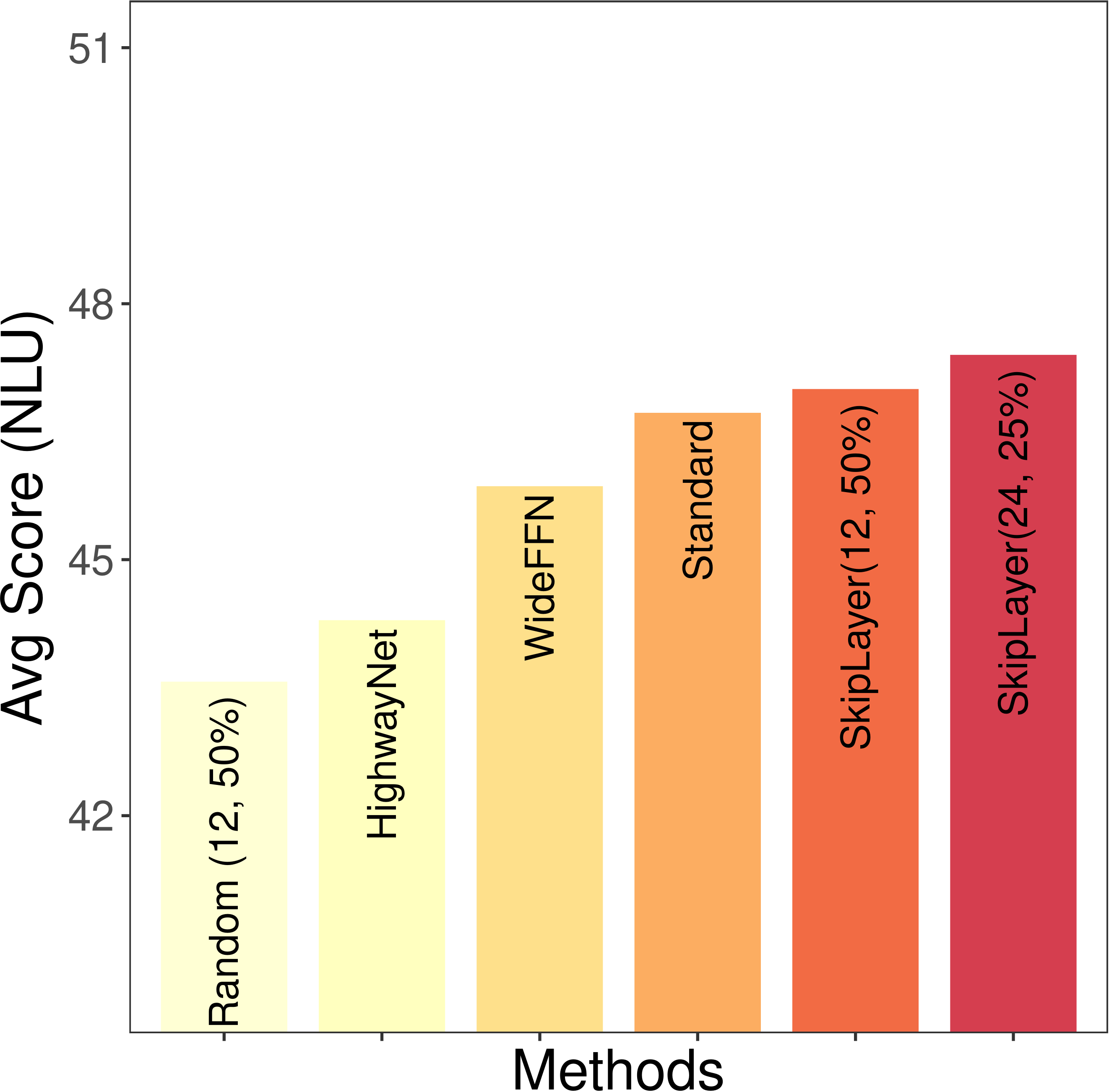}
\caption{Avg. 1-shot (NLU, 6L)}
\end{subfigure}
&
\begin{subfigure}[b]{0.25\textwidth}
\includegraphics[width=\textwidth]{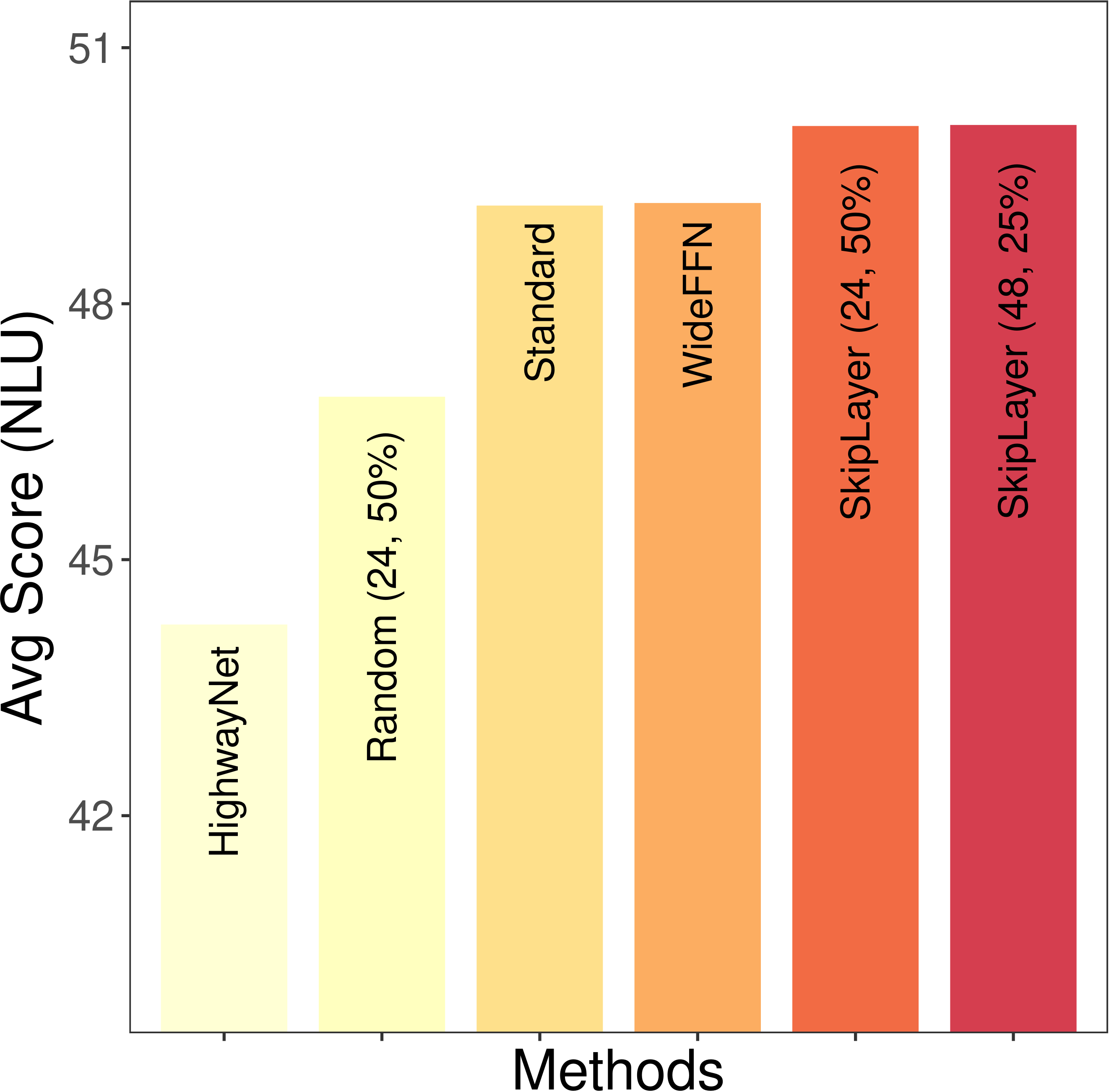}
\caption{Avg. 1-shot (NLU, 12L)}
\end{subfigure}
\end{tabular}
\caption{Average 1-shot NLG and NLU performance of different methods with 6 (a-b) and 12 (c-d) effective number of activated layers, respectively.}
\label{fig:1shot-baselines}
\end{figure*}

\begin{figure}[tb]
\centering
    \renewcommand\tabcolsep{2pt}
\begin{tabular}{cc}
\begin{subfigure}[b]{0.23\textwidth}
\includegraphics[width=\textwidth]{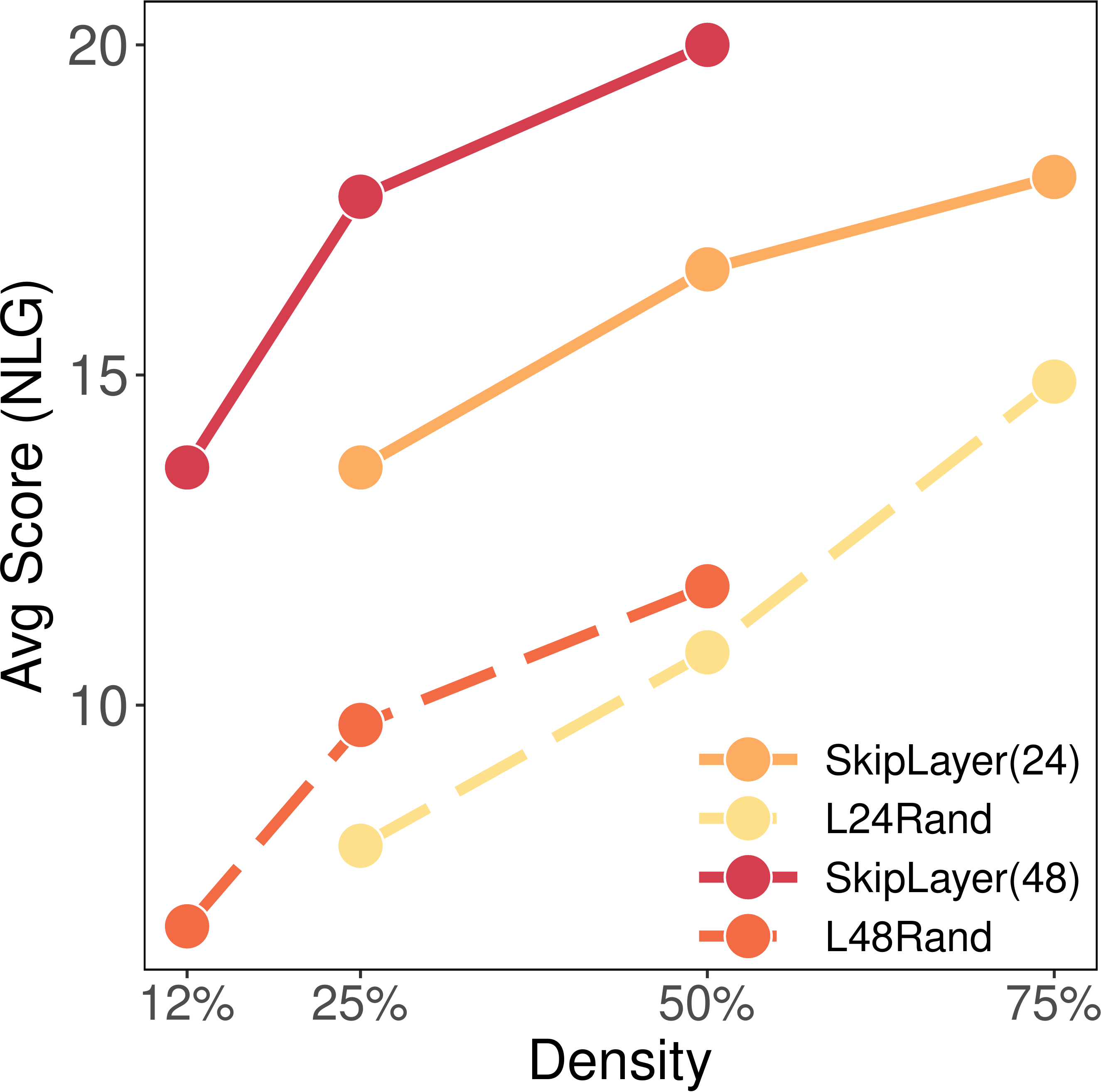}
\caption{Avg. 1-shot (NLG)}
\end{subfigure}
&
\begin{subfigure}[b]{0.23\textwidth}
\includegraphics[width=\textwidth]{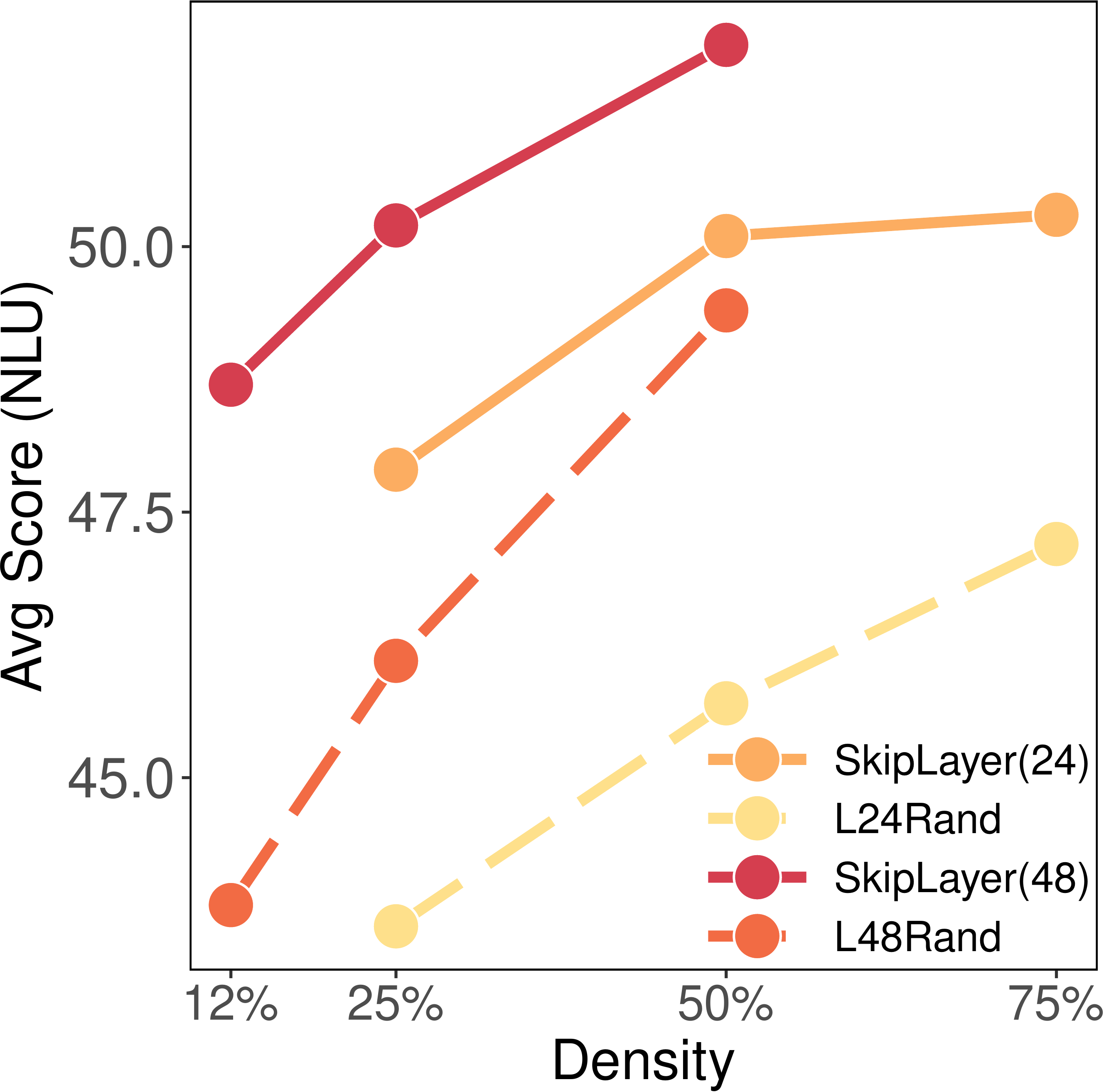}
\caption{Avg. 1-shot (NLU)}
\end{subfigure}
\end{tabular}
\caption{Learned gating has significantly better performance than the respective methods using random gating.}
\label{fig:gating}
\end{figure}

\section{Full Results}
We have conducted comprehensive evaluation of the effectiveness of~\skipl and report the quantitative results compared to different baselines in this section.

\paragraph{How does~\skipl perform in 1-shot learning?}~\skipl allows each input to selectively activate a particular layer depending on the context. By keeping the average number of activated layers constant while increasing the total number of layers of a language model, we expect that the increased model capacity can improve the predictive quality of few-shot learning. Thus, in Figure~\ref{fig:performance}(a-b), we first report the average 1-shot performance of different~\skipl models. In Figure~\ref{fig:performance}(a), the y-axis is the average 1-shot performance of all the NLG tasks, and the x-axis (log-scale) is the compute FLOPs per token during the forward-pass. The black line first shows the performance of the standard baseline models of 6, 12, 24 and 48 layers, respectively. For each baseline model of a particular number of layers, e.g., 6L, we report the performance of the respective~\skipl models of comparable compute FLOPs. For example, the three yellow dots represent the~\skipl models of 12 layers (12L) with 50\% density, 24 layers (24L) with 25\% density, and 48 layers (48L) with 12.5\% density, respectively. Compared to the 6L baseline model, all these three~\skipl models (in yellow) have the same average number of activated layers of six ({\textbf{Eff}}ective number of layers) denoted as Eff06. Similarly, the orange line (Eff12) and red line (Eff24) denote the~\skipl models with 12 and 24 average number of activated layers, accordingly.

In all the cases, it first shows that as the~\skipl models become deeper and sparser (keeping the same activated number of layers), it keeps improving the few-shot learning performance at the modest cost of increased FLOPs per token prediction. For instance, ~\skipl(24L, 25\%) has 51.5\% performance gain at the cost of 18.2\% increased compute FLOPs compared to the 6L baseline. Similarly, ~\skipl(48L, 25\%) and~\skipl(96L, 25\%) have 28\% and 22\% at the cost of 19.2\% and 20\% compared to the 12L and 24L baselines, respectively. Moreover, in Figure~\ref{fig:performance}(a) we can also observe that even though \skipl (48L, 12.5\%) has less compute FLOPs compared to the 12L baseline, it has achieved pretty close 1-shot performance. Similarly, \skipl (96L, 12.5\%) has even better 1-shot performance compared to the 24L baseline by using less compute FLOPs. This verifies that we are able to trade off model capacity for better predictive quality.
Likewise, Figure~\ref{fig:performance}(b) shows similar patterns that increased model capacity can lead to better predictive quality across the NLU tasks while at the cost of modest increased FLOPs.



\paragraph{Does~\skipl decode and train efficiently?} As shown in Figure~\ref{fig:performance}(a-b), deeper and sparser~\skipl models have consistent performance improvement in few-shot learning. We are also interested in studying if they are able to decode and train fast. Figure~\ref{fig:performance}(c) compares the decoding time per token of different models using a single TPU v3 chip. It shows that~\skipl(12, 50\%) has nearly the same per-step decoding time as the baseline 6L. ~\skipl(24, 50\%) also has similar speed as the baseline 12L. ~\skipl(48, 50\%) has 8\% 1-shot performance gain at the cost of 6\% increase in the per-step decoding time compared to the baseline 24L. As the models become deeper, the per-step decoding time also increases. However, we may find some good trade-off between quality and speed. For example,~\skipl(96, 25\%) has achieved 20\% decoding efficiency with a tiny quality loss of only 0.5\% compared to the baseline 48L. Figure~\ref{fig:performance}(d) further shows the training speed on a single TPU v4 of different~\skipl models. In general, the training speed decreases as more FLOPs are used per prediction. Compared to the respective full dense baselines, SkipLayer (24, 50\%) has 18\% speed gains, and SkipLayer (48, 12.5\%) has 3x gains.

\paragraph{How does~\skipl compare to the baselines?} Figure~\ref{fig:1shot-baselines}(a-b) first compare the average 1-shot performance of different methods of effective 6 layers and 12 layers across all the NLG tasks. In Figure~\ref{fig:1shot-baselines}(a) when all models are small, HighwayNet has similar performance as the standard baseline 6L model. They both outperform the Random gating method but marginally underperform the WideFFN method. When all the models become deeper in Figure~\ref{fig:1shot-baselines}(b), the standard baseline 12L scales better than WideFFN, HighwayNet and the Random gating method. However, in both cases, SkipLayer based models have achieved the best performance compared to all the other baselines. HighwayNet performs the worst among all the models we trained, one possible reason could be Highway network is designed for FFN only architectures, it works like a weighted residual branch added to the wrapped layer, which helps the training stability of very deep FFN only networks. However, residual branch is quite normal in today's Transformer models, self-attention and FFN both have residuals. In our implementation, we replace the original residual with the highway network residual which leads to performance degradation.
Figure~\ref{fig:1shot-baselines}(c-d) further compare the average 1-shot performance of different methods across all the NLU tasks. With a similar trend observed before, WideFFN has a scaling performance closer to the standard baseline models of 12 layers, both of which significantly outperform the HighwayNet and the Random gating method, and SkipLayer based models lead to the best performance overall when the models scale up. 

\paragraph{Does the learned gating matter?} The gating function inside a SkipLayer enables each input example to activate a subset of layers of the model based on the context. We have already observed that this flexibility of switching model parameters improve the predictive accuracy of the model when it becomes deeper and sparser. We wonder how much gain the learned routing function can contribute to the predictive performance of the \skipl-based models. We approach this question by comparing the \skipl-based models with the respective Random gating baselines where the gating is not learned by varying the density of the model side-by-side in Figure~\ref{fig:gating}. Because each token activates a layer randomly and independently, Random gating baselines have the similar compute FLOPs per token prediction as the~\skipl-based models given the same model density. However, as shown in Figure~\ref{fig:gating}, for the same model density, the learned gating of ~\skipl-based models performs significantly better than the respective methods using random gating. Even though Random gating baselines also have the flexibility of switching model parameters per token prediction, Figure~\ref{fig:gating} verifies that the learned gating functions are much more effective to improve the prediction accuracy.

\paragraph{How does the skipping behave during decoding?}

\begin{figure}
    \centering
    \includegraphics[clip, trim=50pt 30pt 40pt 30pt, width=0.95\linewidth]{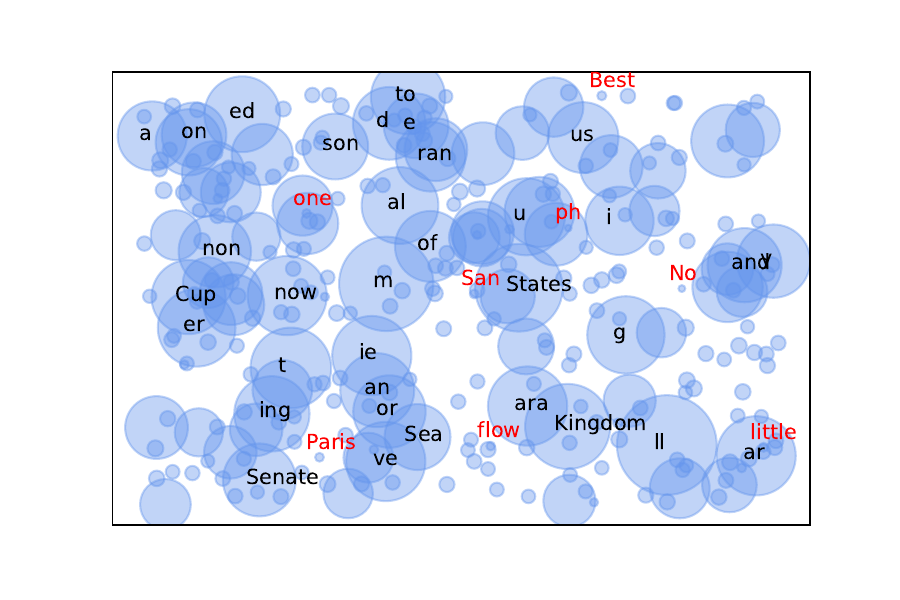}
    \caption{Bubble chart of the skipping behaviour of tokens during greedy decoding on TriviaQA dataset using our \skipl (12L, 50$\%$) model. Each dot represents a token, larger dot size means more layers are skipped. Black/Red texts show some tokens that skip the most/least layers.}
    \label{fig:skip_pattern_bubble}
\end{figure}

To study the skipping behaviour of \skipl model and understand what kind of tokens skip more layers than others during greedy decoding, we collected the skipping statistics of the decoding results of 500 sampled questions in TriviaQA using our \skipl (12L, 50$\%$) model. In Figure \ref{fig:skip_pattern_bubble}, we plotted the bubble chart of 300 frequently used tokens in the decoding results according to their averaged number of skipped layers (One token may have different skipping patterns under different contexts). Larger dots in the figure represents more layers are skipped. We can observe that tokens that skip the most are mainly functional words like ``and", ``to", ``ed" or ``ing". These tokens can be easily inferred from the previous contexts and thus do not need much computation to decode. While tokens that skip less are usually independent words like ``No" or ``Paris". This indeed shows that our \skipl model can successfully identify such tokens and assign proper computation accordingly.

\section{Related Work}
\paragraph{Conditional Computation}~\cite{BengioLC13, BengioBPP15}
is a paradigm where only a subset of model parameters can be activated per input example. Early-exit is one kind of implementation for conditional computation where external classifiers equipped with confidence-based thresholding are used to exit early without going through the whole stack of layers~\cite{WangLCTG17, xin2020deebert, schwartz-etal-2020-right, liu-etal-2020-fastbert, DabreRF20, ElbayadGGA20, tal22}.
Unlike these approaches where the computation is activated for the bottom layers, ~\skipl-based models technically allows each input to explore $2^{L}$ different compute paths for a model with $L$ stacked layers.

An alternative technique is to enable the model to `learn' how to activate its different sub-layers. Due to the discreteness of the activating decisions, soft-approximations and RL-based implementations has been explored in the vision~\cite{srivastava2015highway, WangYDDG18} and NLP~\cite{bapna2020controlling} community. Our approach is closer to the second learning approach but differ in that~\skipl does not require the soft-approximation during the forward pass, giving computational savings not just in inference but also during training. This difference is crucial since pretraining language models is often time consuming and costly.

Additionally, concurrent work such as CODA~\cite{lei2023conditional} and CoLT5~\cite{ainslie2023colt5} have applied similar token selection method to activate Transformer layers. However, these work only apply conditional activation in the encoder layers of an encoder-decoder model such as T5. 


\paragraph{Mixture of Experts} have recently been proposed to improve model efficiency~\cite{shazeer2017outrageously, gross2017hard, lepikhin2020gshard, fedus2021switch, rol21, du2022glam, art21, Lewis2021BASELS, zhou2022mixture, pmlr-v162-rajbhandari22a} by sparsely activating a subset of experts in a MoE layer.

Our approach is orthogonal to MoE models in that an MoE layer can be easily wrapped by the~\skipl for additional efficiency. Moreover,~\skipl can apply conditional computation to both the self-attention and the Feed-Forward (FFN) component of a Transformer layer, whereas MoE models mainly focus on conditionally activating the FFN component in a MoE layer.

\paragraph{Structural Dropout} \cite{tompson2015efficient, ghiasi2018dropblock, dai2019batch, fan2019reducing, zeng2021correlation} randomly drops a group of weights, e.g., a layer~\cite{fan2019reducing}, during training to achieve better generalization and robustness for pruning during inference. However, the amount of computation during inference is still uniform per example. In contrast, ~\skipl-based models learn the skipping patterns from the data which shows better performance than the random skipping baseline, and potentially assign non-uniform amount of computation to each example during inference. 

\section{Conclusions}
We propose a new general method named \skipl for dynamically skipping the execution of arbitrary layers based on the input context using a simple routing algorithm. This method enables heterogeneous computation for tokens at different complexity or importance so that more computation resources can be used for improving the predictive quality of harder tokens. Our model demonstrates significant 1-shot performance improvement across 24 NLP tasks compared to other competitive baselines with only a small extra cost for inference.

\bibliography{gshard, reference}

\bibliographystyle{icml2022}
\clearpage

\end{document}